\documentclass[sigconf]{acmart}
\usepackage{booktabs}
\usepackage{multirow}
\usepackage{graphicx}
\usepackage{tabularx}
\usepackage{booktabs}
\AtBeginDocument{%
  }

\copyrightyear{2026}
\acmYear{2026}
\setcopyright{cc}
\setcctype{by}
\acmConference[MM '26] {Proceedings of the 34th ACM International Conference on Multimedia}{November 10--14, 2026}{Rio de Janeiro, Brazil.}
\acmBooktitle{Proceedings of the 34th ACM International Conference on Multimedia (MM '26), November 10--14, 2026, Rio de Janeiro, Brazil}
\acmISBN{979-8-4007-2213-4/2026/11}
\acmDOI{10.1145/3767308.3836517}

\begin{document}

\title{ID-VTG: Image-Disambiguated Video Temporal Grounding}

\author{Minghang Zheng}
\affiliation{
\institution{Wangxuan Institute of Computer Technology, Peking University}
\city{Beijing}
\country{China}
}
\email{minghang@pku.edu.cn}

\author{Jingli Wei}
\affiliation{
\institution{Wangxuan Institute of Computer Technology, Peking University}
\city{Beijing}
\country{China}
}
\email{jingliwei@stu.pku.edu.cn}

\author{Hongyi Yang}
\affiliation{
\institution{Wangxuan Institute of Computer Technology, Peking University}
\city{Beijing}
\country{China}
}
\email{hyyang25@stu.pku.edu.cn}

\author{Yang Liu}
\authornote{Corresponding author.}
\affiliation{
\institution{Wangxuan Institute of Computer Technology, Peking University}
\institution{State Key Laboratory of General Artificial Intelligence, Peking University}
\city{Beijing}
\country{China}
}
\email{yangliu@pku.edu.cn}

\renewcommand{\shortauthors}{Minghang Zheng et al.}

\begin{abstract}
Video Temporal Grounding (VTG) faces significant challenges when natural language queries must distinguish between multiple events involving visually similar entities, particularly when relying on fine-grained visual attributes that are difficult to describe accurately in words alone. To address this, we introduce Image-Disambiguated Video Temporal Grounding (ID-VTG), a task that leverages multimodal queries combining a reference image and a text description to precisely localize segments where a specific instance performs a described action. To facilitate research, we construct two benchmarks: IDVTG-Gym, focusing on fine-grained, compositionally ordered gymnastics actions with athletes in similar uniforms; and IDVTG-InternVid, an open-world dataset featuring diverse entities (e.g., humans, animals, fictional characters) and significant temporal distractors. Methodologically, we propose the Visually-Guided Disambiguation Aggregation (VGD-Agg) framework based on a dual-branch fast-slow architecture. The fast branch efficiently generates preliminary event proposals, while the slow branch performs fine-grained frame-level matching between video frames and the reference image. We enhance discriminability via two learnable tokens: a Compare Token, which represents hard negatives to probe for the presence of the target instance (as referred to by the query image), and a Depress Value, which represents text-irrelevant events. Proposals that the Compare Token identifies as lacking the target instance are pushed toward the Depress Value, thus easing disambiguation via the text query. Extensive experiments validate our approach, which achieves state-of-the-art results on the proposed benchmarks. Code is available at \url{https://github.com/oceanflowlab/ID-VTG}.

\end{abstract}


%

\maketitle

\section{Introduction}
\label{sec:intro}

\begin{figure}
    \centering
    \includegraphics[width=0.8\linewidth]{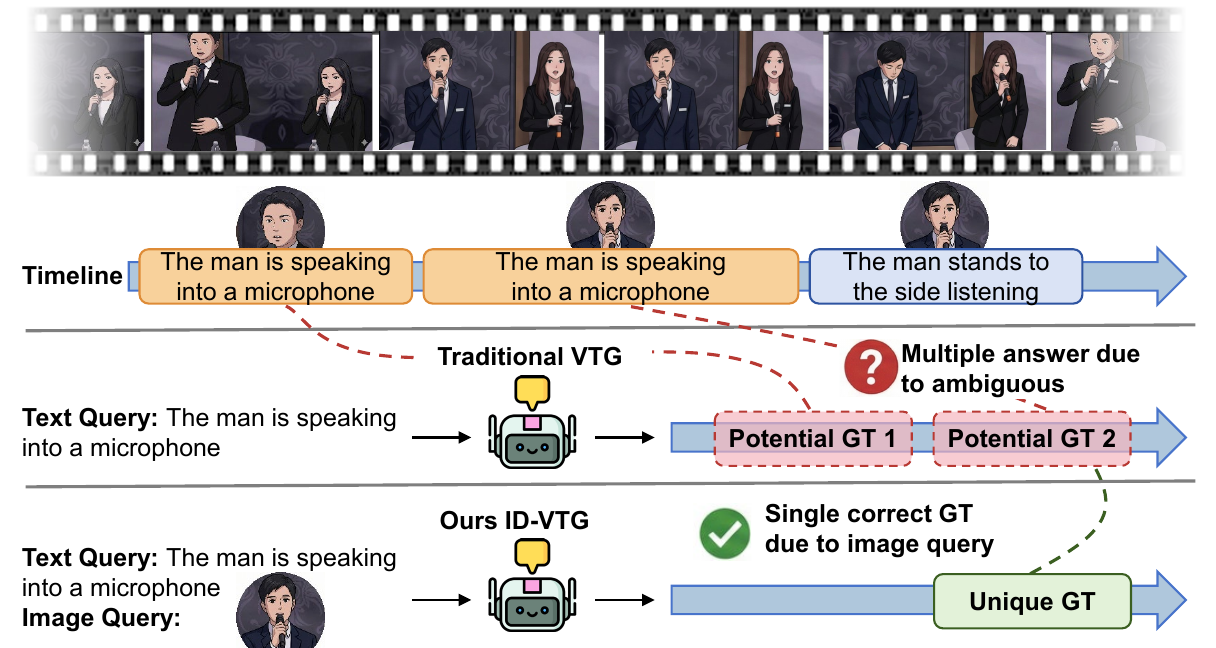}
    \caption{\textbf{Illustration of the proposed Image-Disambiguated Video Temporal Grounding (ID-VTG) task.}}
    \label{fig:teaser}
    \Description{Illustration of the proposed Image-Disambiguated Video Temporal Grounding (ID-VTG) task}
\end{figure}

Video Temporal Grounding (VTG) localizes events in videos from natural language queries. However, existing methods struggle when text cannot distinguish among visually similar entities, especially when differentiation depends on fine-grained attributes (e.g., texture or facial appearance) or unfamiliar subjects that are hard to describe precisely. Such cases lead to ambiguous queries and inaccurate localization. As shown in Figure~\ref{fig:teaser}, when two similar subjects both perform ``the man is speaking into a microphone" at different times, text-only VTG cannot reliably refer to the intended segment. To address this limitation, we introduce Image-Disambiguated Video Temporal Grounding (ID-VTG), a task that supplements text with a reference image. This task directly aligns with critical real-world scenarios where users possess a strong visual prior but find textual descriptions cumbersome and imprecise. Practical applications include intelligent surveillance (locating a specific suspect's actions using a photo), sports broadcasting (tracking a specific athlete), and multimodal retrieval (finding exact products in videos)~\cite{E230419}. By serving as a strict visual constraint, the reference image naturally eliminates semantic ambiguity, enabling precise localization of the intended subject.

\begin{figure*}[t]
    \centering
    \includegraphics[width=0.9\textwidth]{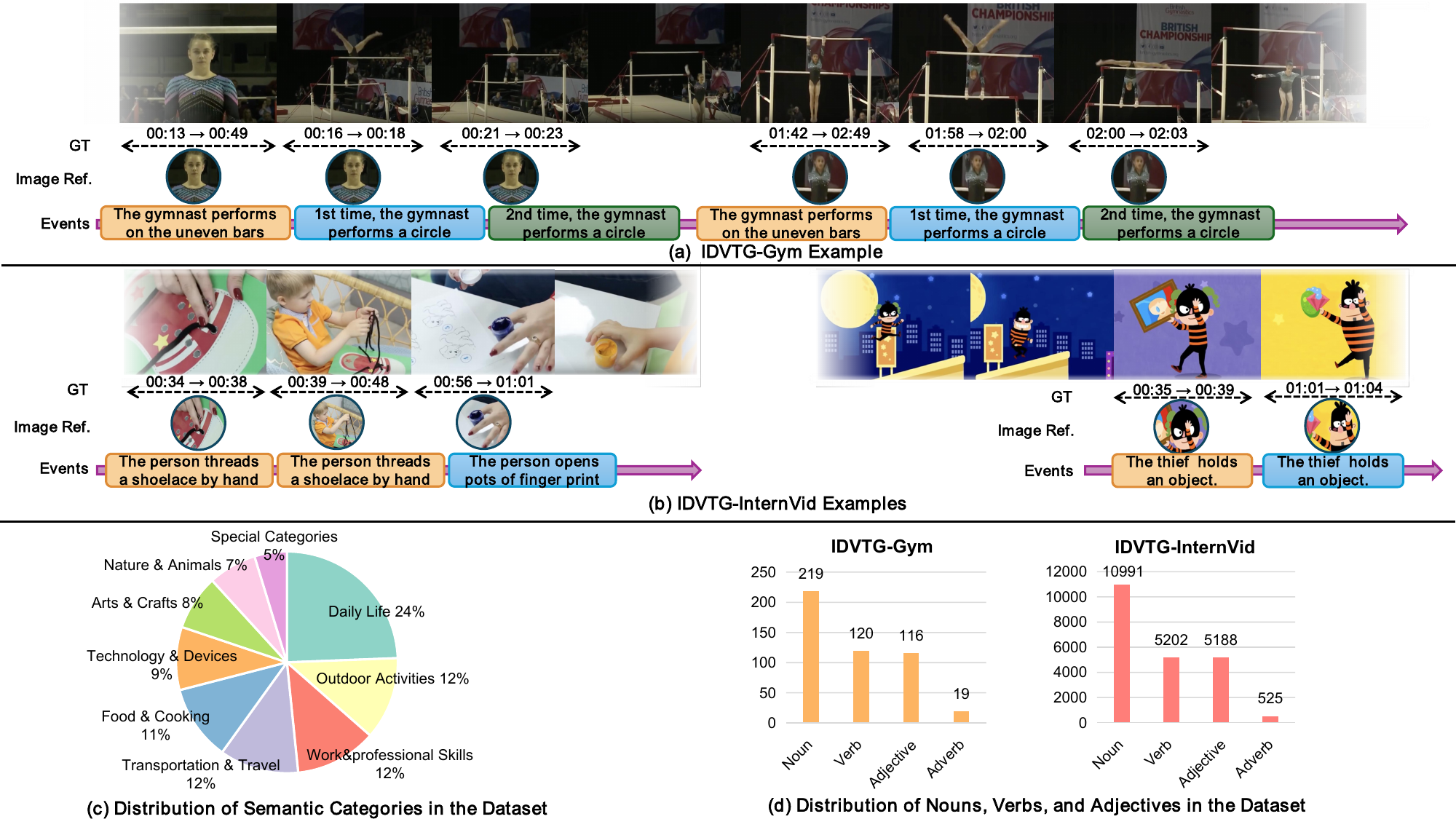}
    \caption{\textbf{Overview of the ID-VTG datasets.}
    (a) IDVTG-Gym focuses on fine-grained, ordered gymnastics actions with visually similar athletes. (b) IDVTG-InternVid focuses on open-world videos with diverse entities and strong temporal distractors.
    (c) The semantic category distribution of IDVTG-InternVid.
    (d) The lexical distribution of IDVTG-Gym and IDVTG-InternVid.}
    \label{fig:teaser_b}
    \Description{Overview of the ID-VTG datasets}
\end{figure*}
To facilitate research on this task, we introduce two comprehensive datasets: IDVTG-Gym and IDVTG-InternVid, both pairing multimodal queries with precise temporal annotations. 
\textit{IDVTG-Gym features fine-grained, compositionally ordered actions.} It is built from gymnastics videos in which athletes wear similar team uniforms, creating substantial visual ambiguity. As seen in Figure~\ref{fig:teaser_b} (a), these reference images are paired with textual queries that range from holistic events (e.g., “The gymnast performs on the uneven bars”) to atomic sub-actions (e.g., “giant circle”). The annotations additionally encode ordinal cues (e.g., “for the second time”), forcing models to reason over action order rather than rely on simple action recognition. \textit{IDVTG-InternVid, in contrast, is designed for open-world settings with diverse content and pronounced distractors.} Figure~\ref{fig:teaser_b}(c) and (d) reveal its coverage of varied video types, including humans, objects, animals, and fictional characters, accompanied by a broader vocabulary.
IDVTG-InternVid also includes strong distractors. As illustrated in Figure~\ref{fig:teaser_b}(b), two different individuals successively perform the action ``threads a shoelace by hand", requiring the model to distinguish subtle, fine-grained differences to disambiguate these highly confusable temporal segments.

To tackle the challenges in ID-VTG, we propose the Visually-Guided Disambiguation Aggregation (VGD-Agg) framework with a dual-branch fast-slow design that balances temporal modeling and fine-grained visual matching: the fast branch efficiently produces preliminary temporal event proposals from high-level video semantics, while the slow branch performs frame-level video-reference matching to disambiguate visually similar instances.
To robustly handle visual ambiguity, we introduce two learnable and video-specific representations: a Compare Token and a Depress Value. The Compare Token acts as a high-similarity hard negative prototype, serving as a threshold to distinguish frames containing the query image from those that do not. Conversely, the Depress Value captures text-irrelevant video events; features of proposals lacking the query image are pushed toward this value to make it easier to differentiate based on the text query. Functionally, a frame is deemed image-relevant only if its affinity to the query image exceeds its similarity to the Compare Token. Accordingly, our Vision-Assisted Disambiguation module enhances relevant proposals by aggregating matched frames, while pulling image-absent proposal features to the Depress Value. This produces clearly separable representations, enabling the subsequent Text-Guided Grounding module to classify proposals and regress boundaries.

Our contributions are: (1) We introduce Image-Disambiguated Video Temporal Grounding (ID-VTG) and two datasets, IDVTG-Gym and IDVTG-InternVid. (2) We propose the VGD-Agg framework to suppress visual distractors and adaptively aggregate features for precise disambiguation-oriented grounding. (3) Extensive experiments show that our method significantly outperforms adapted state-of-the-art VTG baselines, establishing a strong benchmark for multimodal video grounding.

\section{Related Works}

\subsection{Methods for Video Temporal Grounding}
Video temporal grounding aims to localize the segment in a video with a text query. Existing methods are text-centric and fall into two categories:
proposal-based 
approaches~\cite{gao2017tall,zhang2020learning,wang2022negative,zheng-etal-2024-training,Luo_2024_WACV,TRM_2023_AAAI,CPL_2022_CVPR,CNM_2022_AAAI,zheng2025hierarchical,liu2026large}, and proposal-free approaches~\cite{soldan2021vlg,Mu2024SnAG, mun2020local,liu2024towards,xiao2024bridging,SPL_2023_ACL,VMR_2023_CVPR,huang2022emb,lei2021MomentDETR,zheng2026omnivtg,Zheng_2026_ICML,zheng2025weakly}. 
Despite strong performance, these methods rely solely on textual queries, which becomes a bottleneck when descriptions cannot distinguish between visually similar entities. Recent efforts have incorporated visual cues but do not address true multimodal disambiguation. Minotaur~\cite{Khurana2023Minotaur} supports image or text queries individually, but does not support the joint use of both. \citet{zhang2025localizing} explores image-composed queries by reconstructing annotations from QVHighlights, their focus is on semantic modification, where the text describes how the reference image context should be conceptually altered. This leads to their benchmarks lacking ambiguity and being unsuitable for evaluating disambiguation capabilities.
In contrast, our ID-VTG task addresses \textit{referential ambiguity}, using real-world reference images as a strict visual constraint to filter out plausible distractors. To this end, we propose a novel framework that explicitly leverages image queries to disambiguate text-grounded proposals, ensuring precise localization of the target identity.

\subsection{Datasets for Video Temporal Grounding}
Existing datasets~\cite{Krishna2017ActivityNetCaptions,gao2017tall,Regneri2013TACoS,du2025dataset,zhang2020doesexistspatiotemporalvideo,mo2025advancing,tang2021humancentricspatiotemporalvideogrounding} differ in annotation granularity and ambiguity. Datasets such as ActivityNet Captions~\cite{Krishna2017ActivityNetCaptions}, Charades-STA~\cite{gao2017tall}, and TACoS~\cite{Regneri2013TACoS} provide sentence--time interval pairs. However, their queries consist exclusively of text without reference images, and these videos typically contain a single salient actor or dominant event, making the textual query sufficient for localization. Spatio-temporal video grounding (STVG)~\cite{gao2025omniground,gao2026sarl} datasets, such as VidSTG~\cite{zhang2020doesexistspatiotemporalvideo} and HC-STVG~\cite{tang2021humancentricspatiotemporalvideogrounding}, define the task as predicting spatial regions and temporal boundaries as outputs. Although they provide fine-grained spatial bounding boxes as target outputs, their inputs remain text-only. Therefore, their annotation protocols strictly require the target event to be unique within the video, implying that the text description alone is sufficient for localization. Overall, existing benchmarks fall short in evaluating instance-disambiguated VTG due to this lack of hard negatives. To address this, we construct two benchmarks, \textbf{IDVTG-Gym} and \textbf{IDVTG-InternVid}, specifically designed with high-ambiguity scenarios and strong visual distractors for rigorous ID-VTG evaluation with the help of large pretrained models~\cite{gemini-2.5-pro,Qwen3-VL,wei2025poison,shen2025stegoagent,E250336,yang2025vqals,yang2025ar,sa2va}.

\section{Dataset}
\label{sec:dataset}

\begin{figure}
    \centering
    \includegraphics[width=0.9\linewidth]{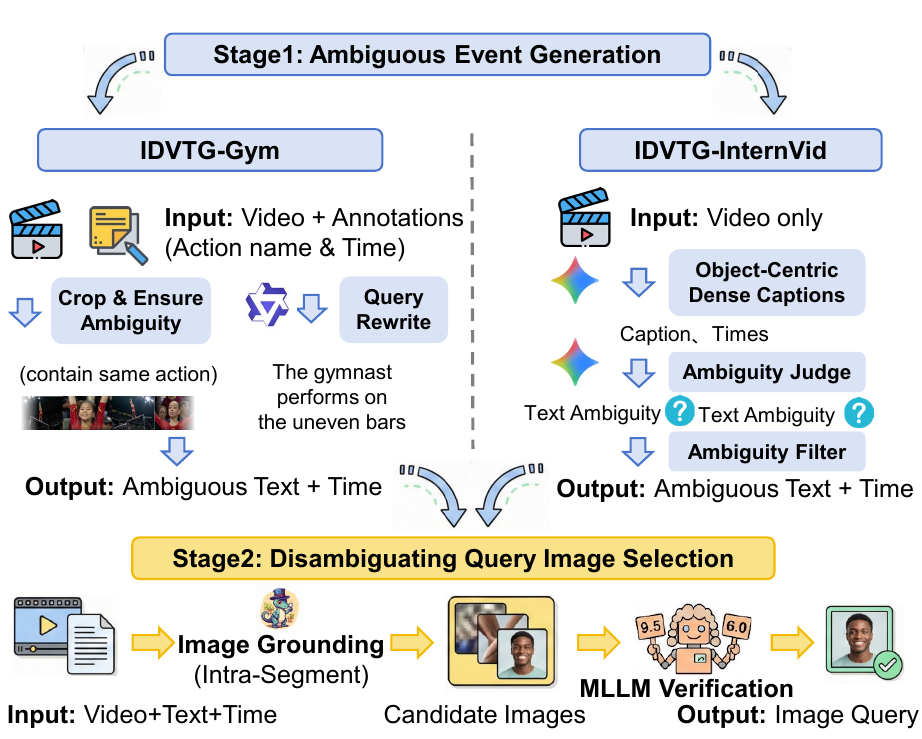}
    \caption{Overview of the ID-VTG dataset construction pipeline.
    \textbf{Stage 1} generates text descriptions and timestamps of ambiguous events for IDVTG-Gym and IDVTG-InternVid.
    \textbf{Stage 2} selects disambiguating image queries by grounding target instances and verifying candidates with an MLLM.
    }
    \label{fig:dataset_pipeline}
    \Description{ID-VTG dataset construction pipeline}
\end{figure}

In support of the ID-VTG task, we construct two datasets, \textbf{IDVTG-Gym} and \textbf{IDVTG-InternVid}, through a unified two-stage pipeline as shown in Fig.~\ref{fig:dataset_pipeline}.
The pipeline first identifies individual \textit{textual ambiguous} instances in each video and assigns textual descriptions with precise temporal boundaries to their events. 
In the second stage, each event is associated with instance-specific spatial regions by grounding the target instance within the corresponding temporal segment, producing \textit{disambiguating visual references} that serve as image queries. 
Through our pipeline, a textual description corresponds to multiple events at different times involving visually similar entities. In such cases, it is difficult to accurately describe fine-grained visual attributes using language alone. Therefore, an image query provides an explicit visual reference, removes ambiguity, and makes precise localization possible.

\subsection{Ambiguous Event Generation}
\label{subsec:text_annotation}

The goal of this stage is to identify temporal segments that cannot be uniquely localized by text alone and output the text query and timestamps. We employ distinct strategies for our IDVTG-Gym and IDVTG-InternVid datasets due to their different annotation information.

\textbf{IDVTG-Gym.}
IDVTG-Gym is constructed from FineGym~\cite{shao2020finegym}, leveraging its hierarchical annotations from event categories (e.g., uneven bars) to atomic sub-actions (e.g., giant circle). Because athletes wear similar uniforms and execute standardized routines, the data naturally exhibits strong visual and textual ambiguity.
We first segment the long videos into minute-level clips and retain only those containing at least two occurrences of the same atomic action label. Since FineGym provides discrete labels, we use an MLLM~\cite{Qwen3-VL} to convert them into free-form text queries and explicitly insert ordinal cues (e.g., the second time) when referring to repeated actions. This enforces ambiguity and requires models to perform sequential temporal reasoning rather than relying on semantic matching.

\textbf{IDVTG-InternVid.}
To increase open-world diversity, we construct the IDVTG-InternVid from InternVid~\cite{wang2024internvidlargescalevideotextdataset} using an automated pipeline with Gemini-2.5-Pro~\cite{gemini-2.5-pro}. First, we instruct Gemini to focus on characters involved in ambiguous events. To ensure \textit{Image Ambiguity} (the visual query alone is insufficient), we require the model to annotate every event performed by the target character with a specific textual query and precise timestamps. This guarantees that the reference image maps to multiple segments, forcing reliance on the text to ground specific action. To ensure \textit{Text Ambiguity} (text query alone is insufficient), for scenarios where different characters perform similar actions, we explicitly instruct the model to generate similar text descriptions. This prevents the text from uniquely identifying the actor, forcing reliance on the reference image. We require Gemini-2.5-Pro to explicitly annotate the ambiguity type and apply a post-filtering step to discard samples that are neither text-ambiguous nor image-ambiguous. After this stage, we further employ another MLLM~\cite{Qwen3-VL} to verify the alignment between each textual description and its corresponding video content, filtering out imperfect samples with weak or inaccurate text-content correspondence. 
To validate the quality of this automatic annotation pipeline and obtain a high-quality test set, we randomly sampled 10\% of the data for manual annotation as a test set. The results show that the automatically labeled data achieved 76.8\% on the R1@0.5 metric compared to the manual annotations.

\subsection{Disambiguating Query Image Selection}
\label{subsec:image_annotation}

Once ambiguous events are defined, the second stage aims to find a query image that uniquely grounds the target event. For each target segment, we sample multiple candidate frames. We employ a pre-trained image grounding model (Sa2VA~\cite{sa2va}) to localize the subject using the text query. A MLLM~\cite{Qwen3-VL} then acts as a verifier, scoring each candidate crop based on subject consistency, image clarity, and the visibility of identity features (e.g., faces). The highest-scoring image is selected as the query. 
As shown in Fig.~\ref{fig:teaser_b}, we present some of our query images. It can be observed that they do not necessarily have to be full-body images and can also depict fictional characters or feature distinctive appearances of fine-grained regions.

\subsection{Evaluation Benchmarks and Statistics}
\label{subsec:test_sets}

We curate specific test sets targeting different distributions. \textbf{Standard Test Sets:} We randomly sample the in-domain test splits for both IDVTG-Gym and IDVTG-InternVid. Because the IDVTG-InternVid dataset’s annotations were automatically generated by Gemini, we re-annotated the test set manually to ensure its high quality. \textbf{Web Dataset:} To test robustness against unseen video sources, we collected and annotated a set of videos from YouTube following the IDVTG-InternVid pipeline. This evaluates the model's ability to generalize to new visual domains.

Tab~\ref{tab:dataset_summary} compares ID-VTG with existing VTG benchmarks. \textbf{IDVTG-Gym} contains 14.7k queries across 204.1 hours of video, focusing on fine-grained actions. \textbf{IDVTG-InternVid} provides a larger scale with 62.1k queries over 302.7 hours, emphasizing open-world vocabulary and diverse video domains. Unlike previous benchmarks, which rely solely on text queries, ID-VTG is the first to integrate \textbf{Text+Image} queries and enforce \textbf{Ambiguity} in the dataset design for video temporal grounding.

\begin{table}[t]
\centering
\caption{Comparison of our ID-VTG dataset with existing VTG datasets.}
\label{tab:dataset_summary}
\resizebox{\linewidth}{!}{
\begin{tabular}{lccccc}
\toprule
Dataset 
& Domain 
& Duration 
& Queries 
& Query Type 
& Ambiguity \\
\midrule
TACoS~\cite{Regneri2013TACoS} 
& Cooking 
& 10.1 h 
& 18.2K 
& Text 
& No \\

Charades-STA~\cite{gao2017tall} 
& Activity 
& 57.1 h 
& 16.1K 
& Text 
& No \\

DiDeMo~\cite{Hendricks2017DiDeMo} 
& Flickr 
& 88.7 h 
& 41.2K 
& Text 
& No \\

QVHighlights~\cite{lei2021MomentDETR} 
& Vlog / News
& 425 h 
& 10.3K 
& Text 
& No \\

ANet-Captions~\cite{Krishna2017ActivityNetCaptions} 
& Activity
& 487.6 h 
& 72.0K 
& Text
& No \\

ICQ-Highlight~\cite{zhang2025localizing} 
& Vlog / News
& 65 h 
& 6.19K 
& Text + Image
& No \\

\midrule

\textbf{IDVTG-Gym (Ours)} 
& Gymnastics
& 204.1 h 
& 14.7K 
& \textbf{Text + Image} 
& \textbf{Yes} \\

\textbf{IDVTG-InternVid (Ours)} 
& Open-world 
& 302.7 h
& 62.1K 
& \textbf{Text + Image} 
& \textbf{Yes} \\

\textbf{ID-VTG (Total)} 
& -- 
& \textbf{506.8 h} 
& \textbf{76.8K} 
& \textbf{Text + Image} 
& \textbf{Yes} \\
\bottomrule
\end{tabular}
}
\end{table}

\section{Method}
\begin{figure*}[t]
    \centering
\includegraphics[width=0.8\linewidth]{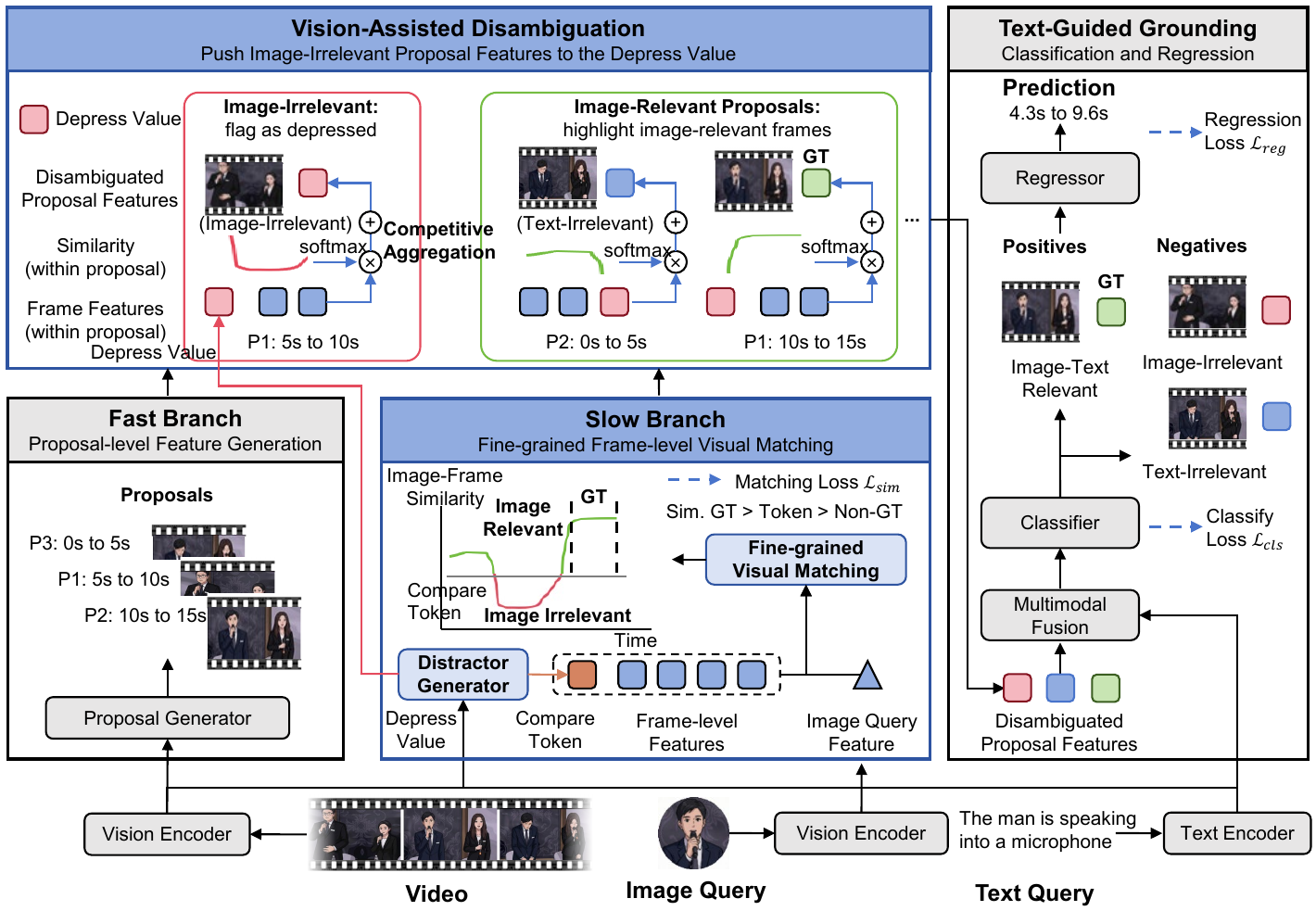}

    \caption{\textbf{Overview of the Visually-Guided Disambiguation Aggregation (VGD-Agg) framework.} The architecture consists of a dual-branch design: (1) The \textbf{Fast Branch} (left) efficiently generates generic, image-agnostic proposal features from the video stream. (2) The \textbf{Slow Branch} (middle) performs fine-grained matching between video frames and the image query to compute similarity scores and learn a video-specific Compare Token. (3) The \textbf{Vision-Assisted Disambiguation} module (top) leverages the Compare Token as a dynamic decision boundary. It acts as a soft gate that suppresses proposal features dominated by visual distractors (\textit{Image-Irrelevant}) while enhancing those containing the target instance (\textit{Image-Relevant}). (4) Finally, the \textbf{Text-Guided Grounding} head (right) aligns the disambiguated features with the text to regress the precise temporal boundaries.}
    \label{fig:method}
    \Description{Overview of the Visually-Guided Disambiguation Aggregation (VGD-Agg) framework.}
\end{figure*}

\noindent \textbf{Problem Definition.} Given a video $\mathcal{V}$ and a multimodal query $\mathcal{Q} = \{ \mathcal{Q}_{text}, \mathcal{Q}_{img} \}$, the goal of ID-VTG is to localize a specific temporal segment $S_g = (t_s, t_e)$, where $t_s$ and $t_e$ denote the start and end timestamps, respectively. The target segment must correspond semantically to the textual description $\mathcal{Q}_{text}$ while visually matching the subject instance depicted in the reference image $\mathcal{Q}_{img}$. 

\subsection{Baseline Revisit}
Our method is built upon SnAG~\cite{Mu2024SnAG}, a representative text-only video temporal grounding framework. Two core modules inherited from the baseline: the \textbf{Fast Branch} for query-agnostic proposal generation and the \textbf{Text-Guided Grounding} module for final grounding.
The Fast Branch efficiently models long-range temporal context and generates candidate event proposals \emph{without} cross-modal alignment, using a Transformer-based multi-scale proposal encoder adapted from ActionFormer~\cite{zhang2022actionformer}.
Given frame-level features, it constructs a hierarchical feature pyramid $\mathbf{Z}$. Each $\mathbf{Z}^{(l)}_i\in\mathbf{Z}^{(l)}$ corresponds to a temporal proposal centered at the $(i\times 2^l)$-th sampled frame with duration $2^l$ sampled frames.
This multi-scale design captures video-wide temporal dependencies and outputs proposal-level representations.
Given proposal features, Text-Guided Grounding fuses proposals and text with a Transformer decoder. The decoding head consists of two parallel multi-layer perceptrons: a classification head and a regression head, predicting proposal confidence $c_i$ and temporal offsets $(o_i^s,o_i^e)$ for boundary refinement, respectively.
Training uses center sampling for positives and optimizes a Focal loss $\mathcal{L}_{cls}$ and a DIoU loss $\mathcal{L}_{reg}$.

\subsection{Overview}

The primary challenge in the ID-VTG lies in resolving the ambiguity where multiple segments match the text description, but only one matches the reference image. Resolving such ambiguity requires incorporating fine-grained visual cues from the reference image to distinguish the target segment. Therefore, we propose the \textbf{Visually-Guided Disambiguation Aggregation (VGD-Agg)} framework. As illustrated in Figure~\ref{fig:method}, our framework is built upon the two baseline components revisited above and introduces two additional modules specifically designed for image-based disambiguation: a \textbf{Slow Branch} and a \textbf{Vision-Assisted Disambiguation} module.

The \textbf{Slow Branch} performs fine-grained frame-level visual matching between video frames and the reference image, enabling vision-sensitive visual discrimination. To robustly differentiate image-relevant frames from visual distractors, we introduce a learnable video-specific Compare Token and its corresponding Depress Value. The Compare Token represents hard negative samples with high similarity to the query image. This design explicitly models challenging distractors and tightens the decision boundary between truly relevant frames and visually similar yet irrelevant ones. The Depress Value represents video events that are irrelevant to the query text. Proposals that do not contain the query image are pushed towards this value, making it easier to differentiate based on the text query.
To explicitly disentangle image-relevant proposals from visual distractors in the feature space, we propose the \textbf{Vision-Assisted Disambiguation} module. This module bridges the two branches via a Softmax-based competition mechanism between video frames and the Compare Token. For image-relevant proposals, the high visual affinity naturally directs the attention weights towards the matching frames, highlighting the target visual content. In contrast, for image-irrelevant distractors (where no frames match the image query), Compare Token dominates the attention distribution, pushing the aggregated proposal features to the Depress Value, providing a clear semantically irrelevant signal to the subsequent text-guided grounding module. Through this mechanism, we obtain \textit{Disambiguated Proposal Features}, where image-relevant targets and distractors are well-separated.

\subsection{Slow Branch}
While the Fast Branch captures generic temporal contexts, the \textbf{Slow Branch} is designed to enable vision-sensitive discrimination by executing fine-grained matching between video frames and the specific image query.

\noindent \textbf{Distractor Generator.} To robustly differentiate image-relevant frames from visual distractors, we propose a distractor generator to generate two complementary video-specific representations: a \textbf{Compare Token} ($\mathbf{t}_{c}$) and a \textbf{Depress Value} ($\mathbf{v}_{d}$). Crucially, we tailor the input queries to the distinct functional role of each representation: $\mathbf{t}_{c}$ serves as a visual affinity baseline and thus requires image-specific guidance, whereas $\mathbf{v}_{d}$ aggregates text-irrelevant visual features to suppress the image-irrelevant proposals and make them easier to differentiate based on the text query. Therefore, we condition the video features $\mathbf{v}$ separately on the visual query $\mathbf{Q}^{\text{visual}}$ and the textual query $\mathbf{Q}^{\text{text}}$ via dedicated Encoder-Decoder architectures: 
\begin{align}
\mathbf{t}_{c} &= \mathrm{Dec}_{tok}(\mathrm{Enc}_{tok}(\mathbf{v}), \mathbf{Q}^{\text{visual}}) \label{eq:line1} \\
\mathbf{v}_{d} &= \mathrm{Dec}_{val}(\mathrm{Enc}_{val}(\mathbf{v}), \mathbf{Q}^{\text{text}}) \label{eq:line3} 
\end{align}
where $\mathrm{Enc}/\mathrm{Dec}$ denote Transformer layers~\cite{vaswani2023attentionneed}, and $\mathbf{v}$ is the video features.
Through this modality-specific conditioning, $\mathbf{t}_{c}$ effectively captures the global visual context to calibrate image-video matching, while $\mathbf{v}_{d}$ encodes text-agnostic visual features to suppress irrelevant proposals. Specifically, $\mathbf{t}_{c}$ is optimized by the Visual Matching Loss described as follows, and $\mathbf{v}_{d}$ is optimized via the Classify Loss $\mathcal{L}_{cls}$ within the Text-Guided Grounding module.

\noindent \textbf{Fine-grained Frame-level Visual Matching.}
To measure the affinity between the video and the reference image, we perform a cross-modal interaction. We first append the Compare Token to the temporal dimension of the original frame-level features, forming an augmented sequence $\mathbf{V}_{in}=[\mathbf{v}; \mathbf{t}_{c}]$. This sequence is then fed into a Transformer~\cite{vaswani2023attentionneed} layer performing cross-attention, where $\mathbf{V}_{in}$ acts as the query, and the visual query features $\mathbf{Q}^{\text{visual}}$ serve as the key and value. This operation injects fine-grained visual cues into both the video frames and the Compare Token $\mathbf{V}_{out}=[\tilde{\mathbf{v}},\tilde{\mathbf{t}}_c]=\mathrm{Dec}(\mathbf{V}_{in}, \mathbf{Q}^{\text{visual}})$, producing the vision-enhanced frame features: $\tilde{\mathbf{v}}$. Finally, a scoring head predicts the similarity scores $s_t$ for each frame and $s_{c}$ for the Compare Token: $s_t, s_{c}=\mathrm{MLP}(\mathbf{V}_{out})$.

\noindent \textbf{Visual Matching Loss.} 
To enforce the role of the Compare Token as a valid discriminator, we introduce a ranking-based loss. Our core objective is to enforce a strict ordinal ranking: $\bar{s}^{\text{gt}} > s_{c} > \bar{s}^{\text{non-gt}}$, which means average similarity within the ground truth segment should exhibit higher visual affinity than the Compare Token, while the Compare Token should score higher than average similarity out of the ground truth segment. In this way, Compare Token serves as a learnable hard negative that probes whether a proposal truly contains the target instance referred to by the query image.
We formulate this objective using a set of hinge losses:s
\begin{align}
\mathcal{L}_{\text{pb}} &= \max(0, m - \bar{s}^{\text{gt}} + s_{c} ) \\
\mathcal{L}_{\text{nb}} &= \max(0, m - s_{c} + \bar{s}^{\text{non-gt}}) \\
\mathcal{L}_{\text{pn}} &= \max(0, 2m - \bar{s}^{\text{gt}} +  \bar{s}^{\text{non-gt}}) \\
\mathcal{L}_{\text{sim}} &= \alpha (\mathcal{L}_{\text{pb}} + \mathcal{L}_{\text{nb}}) + \mathcal{L}_{\text{pn}}
\end{align}
where $\bar{s}^{\text{gt}}$ is the average similarity within the ground truth, $\bar{s}^{\text{non-gt}}$ is the average similarity outside the ground truth, and $m$ is a scalar hyperparameter that specifies the margin enforced by the hinge-loss constraints.

\subsection{Vision-Assisted Disambiguation}
\label{sec:disambiguation}
This module is a bridge that integrates the query-independent proposal features $\mathbf{Z}$ from the Fast Branch with frame features $\tilde{\mathbf{v}}$ from the Slow Branch, achieving feature separability between target events and visual distractors. 

Since the proposals are generated hierarchically, we first align them with the frame-level sequences. For each proposal feature $\mathbf{Z}_i^{(l)}$ at pyramid level $l$, we identify its corresponding temporal receptive field $\mathcal{R}_i^{(l)} = [t_{start}, t_{end}]$ in the frame sequence. We extract the corresponding visual similarity scores $\mathbf{s}_{\mathcal{R}} = [s_t]_{t \in \mathcal{R}_i^{(l)}}$ and the vision-enhanced frame features $\tilde{\mathbf{v}}_{\mathcal{R}} = [\tilde{\mathbf{v}}_t]_{t \in \mathcal{R}_i^{(l)}}$.

\noindent \textbf{Softmax-based Competitive Aggregation.} To implement the adaptive disambiguation, we design a competition mechanism between the specific video frames and the Compare Token. We concatenate the similarity score $s_{c}$ and Depress Value $\mathbf{v}_{d}$ with the extracted frame sequences, forming augmented sets. The aggregated visual feature $\hat{\mathbf{Z}}_i^{(l)}$ is then computed via a softmax-weighted sum:
\begin{equation}
\begin{split}
\mathbf{w}_{attn} = \mathrm{softmax}([\mathbf{s}_{\mathcal{R}}; s_{c}]),\quad
\hat{\mathbf{Z}}_i^{(l)} = \mathbf{w}_{attn}^{\mathrm{T}} \cdot [\tilde{\mathbf{v}}_{\mathcal{R}}; \mathbf{v}_{d}]
\end{split}
\end{equation}
where $[\cdot ; \cdot]$ denotes concatenation along the temporal dimension. The final disambiguated proposal feature $\tilde{\mathbf{Z}}_i^{(l)}= \mathrm{MLP}(\hat{\mathbf{Z}}_i^{(l)} + \mathbf{Z}_i^{(l)})$ is obtained by fusing the vision-aggregated feature with the original proposal feature.

This formulation leverages the Compare Token as an \textit{adaptive baseline} to separate proposals in the feature space.
For \textbf{image-relevant proposals}, the high frame-image affinity within the receptive field \textit{dominates} the attention weights $\mathbf{w}_{attn}$, causing the module to highlight relevant frames.
Conversely, for \textbf{image-irrelevant proposals}, the frame scores fall below the baseline $s_{c}$. Therefore, the attention weights shift significantly towards the Depress Token. This effectively suppresses the misleading visual features of the distractors by replacing them with the Depress Value, providing a clean discriminative signal for the subsequent text-guided grounding module.

\noindent \textbf{Overall Objective.} The model is trained end-to-end by minimizing the weighted sum of the grounding losses and the proposed visual disambiguation similarity loss:
\begin{equation}
    \mathcal{L}_{total} = \mathcal{L}_{cls} + \lambda_{reg} \mathcal{L}_{reg} + \mathcal{L}_{sim}
\end{equation}

\section{Experiment}
\label{sec:experiment}

\subsection{Experimental Setup}
\noindent \textbf{Datasets.} We evaluate on our two benchmarks, \textbf{IDVTG-Gym} and \textbf{IDVTG-InternVid}. 
The queries in the IDVTG-Gym dataset include complete gymnastics events as well as fine-grained gymnastics actions. Therefore, we have divided the test set into two subsets: \textbf{Sub-act.} and \textbf{Holistic}.
IDVTG-InternVid is a large-scale open-world benchmark, evaluated under two settings: 
(1) \textit{In-domain}: testing on its test split; 
(2) \textit{OOD Video}: testing on the \textbf{Web Dataset} with diverse internet videos exhibiting visual distribution shifts.

\noindent \textbf{Evaluation Metrics.} Following standard protocols, we adopt the Recall@$n$, IoU=$m$ ($R^n_m$) and Mean Intersection over Union (mIoU) as our primary metrics. We report results for $n=1$ with $m \in \{0.5, 0.7\}$, as well as the average IoU of the top-1 prediction.

\noindent \textbf{Implementation Details.}
We use pre-trained CLIP (ViT-L/14)~\cite{radford2021clip} to extract frame-level visual and text features with dimension $D=768$. 
The model is trained end-to-end using AdamW~\cite{loshchilov2017adamw} (learning rate $2\times10^{-4}$, weight decay $0.05$), consistent across benchmarks. 
For losses, we set $\alpha=2.0$, $m=1.0$, and $\lambda_{reg}=2.0$.

\subsection{Main Results}

\begin{table}[t]
\caption{Performance comparison on the IDVTG-Gym dataset.}
\label{tab:gym_results}
\centering
\resizebox{\linewidth}{!}{%
\begin{tabular}{lcccccccccc}
\toprule
\multirow{2}{*}{Method}
& \multicolumn{3}{c}{Overall}
& \multicolumn{3}{c}{Sub-act.}
& \multicolumn{3}{c}{Holistic} \\
\cmidrule(lr){2-4}
\cmidrule(lr){5-7}
\cmidrule(lr){8-10}
& $R_{0.5}^{1}$ & $R_{0.7}^{1}$ & mIoU
& $R_{0.5}^{1}$ & $R_{0.7}^{1}$ & mIoU
& $R_{0.5}^{1}$ & $R_{0.7}^{1}$ & mIoU \\
\midrule
RaTSG~\cite{dong2024temporal}
& 18.45 & 15.99 & 16.96
& 6.64 & 3.98 & 5.54
& 26.28 & 23.50 & 27.64 \\

UVCOM~\cite{uvcom}
& 18.07 & 15.00 & 17.32
& 5.20 & 2.43 & 6.86
& 25.07 & 22.54 & 22.99 \\

CG-DETR~\cite{moon2024correlationguidedquerydependencycalibrationvideo}
& 27.96 & 23.36 & 26.99
& 6.64 & 2.77 & 10.14
& 44.47 & 39.86 & 40.29 \\

ICQ~\cite{zhang2025localizing}
& 46.12 & 42.24 & 40.24
& 43.92 & 35.73 & 34.10
& 47.87 & 47.17 & 45.06 \\

SnAG~\cite{Mu2024SnAG}
& 53.64 & 48.82 & 46.73
& 51.66 & 41.37 & 40.32
& 54.83 & 54.05 & 51.46 \\

\textbf{VGD-Agg (Ours)}
& \textbf{61.83} & \textbf{56.53} & \textbf{54.17}
& \textbf{55.53} & \textbf{44.91} & \textbf{43.29}
& \textbf{66.58} & \textbf{65.36} & \textbf{62.53} \\
\bottomrule
\end{tabular}%
}
\end{table}

\begin{table}[t]
\caption{Performance comparison on the IDVTG-InternVid and Web datasets.}
\label{tab:merged_internvid_web}
\centering
\scalebox{0.85}{
\begin{tabular}{lccccccc}
\toprule
\multirow{2}{*}{Method}
& \multicolumn{3}{c}{IDVTG-InternVid}
& \multicolumn{3}{c}{Web} \\
\cmidrule(lr){2-4}
\cmidrule(lr){5-7}
& $R_{0.5}^{1}$ & $R_{0.7}^{1}$ & mIoU
& $R_{0.5}^{1}$ & $R_{0.7}^{1}$ & mIoU \\
\midrule
RaTSG~\cite{dong2024temporal}
& 30.58 & 19.71 & 30.91
& 7.96 & 5.79 & 8.53 \\

UVCOM~\cite{uvcom}
& 30.11 & 18.89 & 30.07
& 7.70 & 5.62 & 8.10 \\

CG-DETR~\cite{moon2024correlationguidedquerydependencycalibrationvideo}
& 44.81 & 31.62 & 43.80
& 13.86 & 5.91 & 18604 \\

ICQ~\cite{zhang2025localizing}
& 31.04 & 25.39 & 29.90
& 7.84 & 4.53 & 8.31 \\

SnAG~\cite{Mu2024SnAG}
& 44.26 & 35.46 & 41.70
& 16.48 & 9.06 & 16.53 \\

\textbf{VGD-Agg (Ours)}
& \textbf{51.21} & \textbf{41.24} & \textbf{48.30}
& \textbf{21.99} & \textbf{13.25} & \textbf{22.34} \\
\bottomrule
\end{tabular}%
}
\end{table}

\noindent \textbf{Baselines.}
We compare VGD-Agg with five representative VTG methods: RaTSG~\cite{dong2024temporal}, UVCOM~\cite{uvcom}, CG-DETR~\cite{moon2024correlationguidedquerydependencycalibrationvideo}, SnAG~\cite{Mu2024SnAG}, and ICQ~\cite{zhang2025localizing}. 
The first four baselines cover both proposal-based and transformer-based paradigms. For adaptation, we follow the common practice in mainstream MLLMs by employing Attention-Fusion. Specifically, image queries are encoded by CLIP and projected into the textual embedding space, then concatenated with text tokens along the sequence dimension. The resulting unified multimodal sequence is processed by the model’s original text encoder, where cross-modal interaction is achieved implicitly through attention. 
For ICQ, we implement its MQ-Sum strategy, which leverages an MLLM~\cite{Qwen3-VL} to jointly summarize the reference image and the textual query into a distilled description, which then serves as the final query for the grounding backbone.

\noindent \textbf{Performance on IDVTG-Gym (Table~\ref{tab:gym_results}).}
VGD-Agg achieves the best performance across all splits, demonstrating its ability to leverage visual guidance for resolving complex ambiguities. Specifically, in the \textbf{Sub-act} split, our model effectively utilizes the reference image as a precise anchor to distinguish fine-grained, atomic movements performed by similarly dressed athletes. Furthermore, on the \textbf{Holistic} split, the method exhibits robust capability in capturing long-range event dependencies, accurately grounding complete routines despite their extended temporal duration.

\noindent \textbf{Performance on IDVTG-InternVid and OOD Web Dataset (Tables~\ref{tab:merged_internvid_web}).} IDVTG-InternVid represents a challenging open-world setting with diverse video content and free-form multimodal queries. VGD-Agg achieves superior performance, significantly outperforming strong baselines and demonstrating that our method effectively scales beyond constrained domains. To further assess robustness, we evaluate the model trained on IDVTG-InternVid directly on the OOD Web dataset without fine-tuning. Despite the domain shift, VGD-Agg maintains a clear lead, indicating that it learns transferable visual-semantic correspondences rather than overfitting to dataset-specific biases. These results validate that explicitly leveraging reference images enables the model to accurately distinguish target instances amid complex, unconstrained entities and actions, and to generalize effectively to unseen data distributions.

\subsection{Ablation Studies}
We conduct comprehensive ablation studies on IDVTG-Gym to validate the contributions of each component.

\begin{table}[t]
\caption{Ablation on branch-level architecture.}
\label{tab:ablation_branch}
\centering
\scalebox{0.85}{
\begin{tabular}{lccc}
\toprule
Model Variant & $R_{0.5}^{1}$ & $R_{0.7}^{1}$ & mIoU \\
\midrule
w/o Slow Branch & 53.64 & 48.82 & 46.73 \\
w/o Fast Branch & 59.41 & 54.40 & 51.32 \\
w/o Aggregation & 61.12 & 56.76 & 53.50 \\
\textbf{VGD-Agg (Full)} & \textbf{61.83} & \textbf{56.53} & \textbf{54.17} \\
\bottomrule
\end{tabular}%
}
\end{table}

\begin{table}[t]
\caption{Ablation on compare token and depress value.}
\label{tab:ablation_bg}
\centering
\scalebox{0.85}{
\begin{tabular}{lccc}
\toprule
Model Variant & $R_{0.5}^{1}$ & $R_{0.7}^{1}$ & mIoU \\
\midrule
w/o Compare Token & 55.63 & 50.76 & 48.49 \\
Global Learnable Token & 60.36 & 55.53 & 52.62 \\
w/o Similarity Loss & 60.50 & 55.58 & 52.67 \\
w/o Image in Compare Token Generation & 60.64 & 56.20 & 53.26 \\
w/o Text in Depress Value Generation & 59.34 & 54.26 & 51.94 \\
\textbf{VGD-Agg (Full)} & \textbf{61.83} & \textbf{56.53} & \textbf{54.17} \\
\bottomrule
\end{tabular}%
}
\end{table}

\begin{table}[t]
\caption{Ablation on different input fusion methods.}
\label{tab:ablation_modal_vgd}
\centering
\scalebox{0.85}{
\begin{tabular}{lccccc}
\toprule
Input Modality & $R_{0.5}^{1}$ & $R_{0.7}^{1}$ & mIoU \\
\midrule
Text-only & 46.78 & 41.96 & 40.74 \\
Attention Fusion & 53.64 & 48.82 & 46.73 \\
BLIP-2 Fusion~\cite{blip2} & 54.07 & 48.86 & 46.75 \\
ICQ Fusion~\cite{zhang2025localizing} & 46.12 & 42.24 & 40.24 \\
\textbf{Text + Image + VGD-Agg (Ours)} & \textbf{61.83} & \textbf{56.53} & \textbf{54.17} \\
\bottomrule
\end{tabular}%
}
\end{table}

\noindent \textbf{Impact of Branch-Level Architecture (Table~\ref{tab:ablation_branch}).} The Slow Branch dominates performance by capturing spatial semantics, while the Fast Branch adds essential temporal context; removing either significantly degrades mIoU. Notably, the full VGD-Agg surpasses the ``w/o Aggregation" baseline, demonstrating our mechanism's ability to effectively synergize spatial and temporal cues.

\noindent \textbf{Effectiveness of Compare Token and Depress Value (Table~\ref{tab:ablation_bg}).}
Removing the token and replacing the softmax competition with a plain softmax over frame similarities (\textbf{w/o Compare Token}) causes a sharp performance drop, proving the necessity of an adaptive baseline to suppress distractors. Furthermore, using a \textbf{Global Learnable Token} or removing the auxiliary supervision (\textbf{w/o Similarity Loss}) leads to inferior results, highlighting that the Compare Token must be video-specific and explicitly optimized as a decision boundary. 
Finally, generating the Compare Token without image context (\textbf{w/o Image}) or generating the Depress Value without text context (\textbf{w/o Text}) also degrades accuracy, indicating that the Compare Token relies on image queries to determine if a video frame includes the image query, while the Depress Value is based on text queries, reflecting video events not related to the text.

\noindent \textbf{Ablation on Input Modalities and Fusion Mechanisms (Table~\ref{tab:ablation_modal_vgd}).}
We conduct ablation studies to examine the impact of image-conditioned VTG and evaluate the effectiveness of the proposed VGD-Agg module. Incorporating visual features consistently improves over the text-only baseline, indicating that reference images provide complementary cues that help resolve semantic ambiguity. We compare several fusion strategies: \textbf{Attention Fusion}, used in baseline adaptation, already delivers substantial gains by explicitly introducing visual guidance. \textbf{BLIP-2 Fusion}, which leverages a pretrained BLIP-2 backbone to combine text and image features, provides only marginal improvements, suggesting that general-purpose multimodal pretraining does not guarantee the fine-grained alignment needed for temporal grounding. \textbf{ICQ Fusion}, which converts image content into textual descriptions using an MLLM, underperforms, highlighting the information loss introduced by text-based compression. Finally, our full \textbf{VGD-Agg} model achieves the best performance, confirming that simply adding images is insufficient; effective image-conditioned grounding requires a dedicated fusion mechanism that aligns and aggregates visual cues with video features in a task-specific manner.

\begin{table}[t]
\caption{Ablation study on different ambiguity types on IDVTG-InternVid.}
\label{tab:ablation_ambiguity}
\centering
\resizebox{\linewidth}{!}{%
\begin{tabular}{lccccccccc}
\toprule
Method
& \multicolumn{3}{c}{Full Amb.}
& \multicolumn{3}{c}{Text Amb. Only}
& \multicolumn{3}{c}{Img Amb. Only} \\
\cmidrule(lr){2-4}
\cmidrule(lr){5-7}
\cmidrule(lr){8-10}
& $R_{0.5}^{1}$ & $R_{0.7}^{1}$ & mIoU
& $R_{0.5}^{1}$ & $R_{0.7}^{1}$ & mIoU
& $R_{0.5}^{1}$ & $R_{0.7}^{1}$ & mIoU \\
\midrule
RaTSG~\cite{dong2024temporal}
& 19.61 & 14.96 & 20.87
& 23.47 & 20.88 & 24.15
& 26.95 & 21.18 &  27.03\\

UVCOM~\cite{uvcom}
& 19.49 & 14.79 & 20.55
& 23.13 & 19.97 & 23.67
& 26.73 & 20.89 & 26.01 \\

CG-DETR~\cite{moon2024correlationguidedquerydependencycalibrationvideo}
& 30.29 & 19.61 & 31.06
& 37.88 & 26.67 & 38.60
& 37.26 & 23.83 & 37.88 \\

ICQ~\cite{zhang2025localizing}
& 20.54 & 15.77 & 20.50
& 25.70 & 21.45 & 25.15
& 27.48 & 21.46 & 27.29 \\

SnAG~\cite{Mu2024SnAG}
& 30.39 & 23.34 & 30.34
& 41.94 & 34.24 & 39.59
& 38.19 & 27.53 &36.40  \\

\textbf{VGD-Agg (Ours)}
& \textbf{37.34} & \textbf{28.42} & \textbf{36.69}
& \textbf{53.27} & \textbf{43.52} & \textbf{49.97}
& \textbf{41.12} & \textbf{30.52} & \textbf{40.12} \\
\bottomrule
\end{tabular}%
}
\end{table}

\begin{table}[t]
\caption{Ablations on the robustness to image queries on IDVTG-InternVid.}
\label{tab:robustness}
\centering
\resizebox{\linewidth}{!}{%
\begin{tabular}{lccccccccc}
\toprule
\multirow{2}{*}{Method}
& \multicolumn{3}{c}{Clean}
& \multicolumn{3}{c}{Brightness}
& \multicolumn{3}{c}{Low Resolution}\\
\cmidrule(lr){2-4}
\cmidrule(lr){5-7}
\cmidrule(lr){8-10}
& $R_{0.5}^{1}$ & $R_{0.7}^{1}$ & mIoU
& $R_{0.5}^{1}$ & $R_{0.7}^{1}$ & mIoU
& $R_{0.5}^{1}$ & $R_{0.7}^{1}$ & mIoU \\
\midrule
RaTSG~\cite{dong2024temporal}
& 30.58 & 19.71 & 30.91
& 29.04 & 18.87 & 28.99
& 29.98 & 18.35 & 29.31 \\

UVCOM~\cite{uvcom}
& 30.11 & 18.89 & 30.07
& 28.76 & 18.16 & 28.90
& 29.83 & 18.51 & 29.84 \\

CG-DETR~\cite{moon2024correlationguidedquerydependencycalibrationvideo}
& 44.81 & 31.62 & 43.80
& 44.35 & 31.60 & 43.36
& 44.31 & 31.54 & 43.27\\

ICQ~\cite{zhang2025localizing}
& 31.04 & 25.39 & 29.90
& 29.00 & 23.29 & 28.02
& 30.77 & 24.66 & 29.54 \\

SnAG~\cite{Mu2024SnAG}
& 44.26 & 35.46 & 41.70
& 43.57 & 35.05 & 41.18
& 43.48 & 35.13 & 41.15 \\

\textbf{Ours}
& \textbf{51.21} & \textbf{41.24} & \textbf{48.30}
& \textbf{49.99} & \textbf{39.79} & \textbf{47.20}
& \textbf{49.10} & \textbf{39.33} & \textbf{46.52} \\
\bottomrule
\end{tabular}%
}
\end{table}

\noindent \textbf{Analysis on Different Ambiguity Types (Table~\ref{tab:ablation_ambiguity}).} To analyze the model's disambiguation ability under different conditions, we categorize the ambiguity in the IDVTG-InternVid dataset into two types: Text Ambiguity and Image Ambiguity. Text Ambiguity refers to scenarios where the textual description applies to multiple visual instances (e.g., different characters performing the same action), necessitating the reference image to identify the specific target. Image Ambiguity occurs when the reference subject appears in multiple temporal segments performing different actions, requiring the textual query to localize the intended event.
As Table~\ref{tab:ablation_ambiguity} shows, VGD-Agg performs robustly in both cases, with larger gains under Text Ambiguity. This demonstrates that the Compare Token and Depress Value effectively suppress visually similar distractors. Furthermore, the strong results on Image Ambiguity indicate that our method retains the precise temporal grounding capability needed to distinguish different events involving the same subject, without over-relying on visual matching alone.

\noindent \textbf{Robustness to Image Queries (Table~\ref{tab:robustness}).}
In real-world scenarios, user-provided reference images often suffer from unpredictable variations in quality; therefore, it is essential to evaluate the model's stability against such degradations. We evaluate model robustness against baselines under three image conditions: Clean, Brightness, and Low Resolution. The Clean setting uses manually annotated high-quality images. Brightness varies the brightness of the reference images, while Low Resolution downsamples the image by a factor of 4 in pixel count. As shown in Table~\ref{tab:robustness}, our method consistently achieves the best performance across all settings. Under Brightness and Low Resolution, all methods remain relatively stable, indicating a certain degree of robustness to brightness variations and spatial degradation. Our method still achieves the best results in both cases, showing that it can preserve reliable visual grounding under degraded image quality.

\subsection{Qualitative Results}
\begin{figure}
    \centering
    \includegraphics[width=\linewidth]{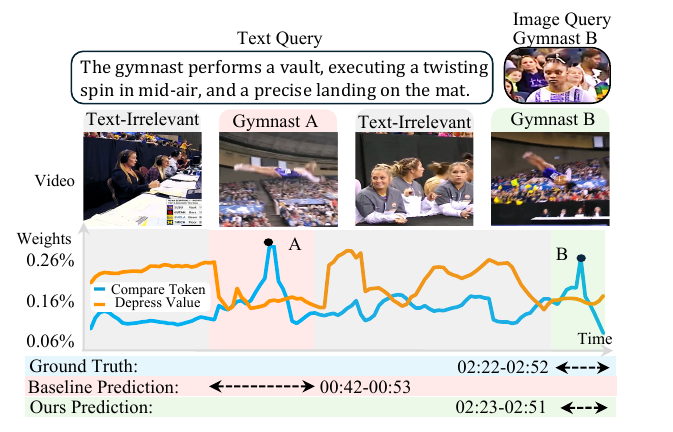}
    \caption{\textbf{Visualization of the learned attention weights.}}
    \Description{qualitative results}
    \label{fig:qual}
\end{figure}
To further validate the effectiveness of our \textbf{Vision-Assisted Disambiguation} module, we visualize the attention weights of the \textit{Compare Token} and \textit{Depress Value} in Figure \ref{fig:qual}. We consider a challenging scenario where two gymnasts perform the same vault activity, with one serving as the visual distractor and the other as the target. The attention weights of the \textit{Compare Token} show clear responses to the high-motion regions of both gymnasts, suggesting that it captures visually ambiguous areas and constructs informative hard negative prototypes for target discrimination. In contrast, the \textit{Depress Value} mainly focuses on text-irrelevant segments, such as the judges in the background, indicating its ability to identify and suppress irrelevant visual content. Together, these visualizations demonstrate that our module effectively distinguishes target-related cues from both hard visual distractors and background noise, thereby facilitating more accurate text-guided grounding.
\section{Conclusion}
In this work, we study Image Disambiguated Video Temporal Grounding (ID-VTG), a setting where text alone is insufficient to resolve multi-instance ambiguity and visual cues become essential for identifying the intended event. We introduce two ID-VTG datasets specifically designed to expose such ambiguities and propose a Visually-Guided Disambiguation Aggregation framework that leverages the reference image to highlight visually consistent regions while suppressing distractors. Experiments validate the effectiveness of our method and highlight the importance of visual disambiguation for reliable temporal grounding in real-world videos.

\begin{acks}
This work was supported by the grants from the National Natural Science Foundation of China 62372014, Beijing Nova Program and ZTE Industry-University-Institute Cooperation Funds under Grant No.IA20240723020-PO0015.
\end{acks}

\bibliographystyle{ACM-Reference-Format}
\balance
\bibliography{acmart}

@inproceedings{CNM_2022_AAAI,
 author = {Zheng, Minghang and Huang, Yanjie and Chen, Qingchao and Liu, Yang},
 booktitle = {Proceedings of the AAAI Conference on Artificial Intelligence},
 title = {Weakly Supervised Video Moment Localization with Contrastive Negative Sample Mining},
 year = {2022}
}

@misc{tang2021humancentricspatiotemporalvideogrounding,
      title={Human-centric Spatio-Temporal Video Grounding With Visual Transformers}, 
      author={Zongheng Tang and Yue Liao and Si Liu and Guanbin Li and Xiaojie Jin and Hongxu Jiang and Qian Yu and Dong Xu},
      year={2021},
      eprint={2011.05049},
      archivePrefix={arXiv},
      primaryClass={cs.CV},
      url={https://arxiv.org/abs/2011.05049}, 
}

@misc{zhang2020doesexistspatiotemporalvideo,
      title={Where Does It Exist: Spatio-Temporal Video Grounding for Multi-Form Sentences}, 
      author={Zhu Zhang and Zhou Zhao and Yang Zhao and Qi Wang and Huasheng Liu and Lianli Gao},
      year={2020},
      eprint={2001.06891},
      archivePrefix={arXiv},
      primaryClass={cs.CV},
      url={https://arxiv.org/abs/2001.06891}, 
}

@inproceedings{CPL_2022_CVPR,
 author = {Zheng, Minghang and Huang, Yanjie and Chen, Qingchao and Peng, Yuxin and Liu, Yang},
 booktitle = {Proceedings of the IEEE/CVF Conference on Computer Vision and Pattern Recognition (CVPR)},
 title = {Weakly Supervised Temporal Sentence Grounding with Gaussian-based Contrastive Proposal Learning},
 year = {2022}
}

@inproceedings{gao2017tall,
 author = {Gao, Jiyang and Sun, Chen and Yang, Zhenheng and Nevatia, Ram},
 booktitle = {Proceedings of the IEEE international conference on computer vision},
 pages = {5267--5275},
 title = {Tall: Temporal activity localization via language query},
 year = {2017}
}

@inproceedings{Luo_2024_WACV,
 author = {Luo, Dezhao and Huang, Jiabo and Gong, Shaogang and Jin, Hailin and Liu, Yang},
 booktitle = {Proceedings of the IEEE/CVF Winter Conference on Applications of Computer Vision (WACV)},
 month = {January},
 pages = {5464-5473},
 title = {Zero-Shot Video Moment Retrieval From Frozen Vision-Language Models},
 year = {2024}
}

@inproceedings{TRM_2023_AAAI,
 author = {Minghang Zheng and Sizhe Li and Qingchao Chen and Yuxin Peng and Yang Liu},
 booktitle = {Proceedings of the AAAI Conference on Artificial Intelligence},
 title = {Phrase-level Temporal Relationship Mining for Temporal Sentence Localization},
 year = {2023}
}

@inproceedings{wang2022negative,
 author = {Wang, Zhenzhi and Wang, Limin and Wu, Tao and Li, Tianhao and Wu, Gangshan},
 booktitle = {Proceedings of the AAAI Conference on Artificial Intelligence},
 pages = {2613--2623},
 title = {Negative sample matters: A renaissance of metric learning for temporal grounding},
 volume = {36},
 year = {2022}
}

@inproceedings{zhang2020learning,
 author = {Zhang, Songyang and Peng, Houwen and Fu, Jianlong and Luo, Jiebo},
 booktitle = {Proceedings of the AAAI Conference on Artificial Intelligence},
 pages = {12870--12877},
 title = {Learning 2d temporal adjacent networks for moment localization with natural language},
 volume = {34},
 year = {2020}
}

@inproceedings{zheng-etal-2024-training,
 author = {Zheng, Minghang  and
Cai, Xinhao  and
Chen, Qingchao  and
Peng, Yuxin  and
Liu, Yang},
 booktitle = {Proceedings of the European Conference on Computer Vision (ECCV)},
 title = {Training-free Video Temporal Grounding usingLarge-scale Pre-trained Models},
 year = {2024}
}

@inproceedings{soldan2021vlg,
  title={VLG-Net: Video-language graph matching network for video grounding},
  author={Soldan, Mattia and Xu, Mengmeng and Qu, Sisi and Tegner, Jesper and Ghanem, Bernard},
  booktitle={Proceedings of the IEEE/CVF International Conference on Computer Vision},
  pages={3224--3234},
  year={2021}
}

@inproceedings{lei2021MomentDETR,
 author = {Lei, Jie and Berg, Tamara L and Bansal, Mohit},
 booktitle = {Advances in Neural Information Processing Systems},
 editor = {M. Ranzato and A. Beygelzimer and Y. Dauphin and P.S. Liang and J. Wortman Vaughan},
 pages = {11846--11858},
 publisher = {Curran Associates, Inc.},
 title = {Detecting Moments and Highlights in Videos via Natural Language Queries},
 url = {https://proceedings.neurips.cc/paper_files/paper/2021/file/62e0973455fd26eb03e91d5741a4a3bb-Paper.pdf},
 volume = {34},
 year = {2021}
}

@inproceedings{Mu2024SnAG,
  author    = {Fangzhou Mu and Sicheng Mo and Yin Li},
  title     = {SnAG: Scalable and Accurate Video Grounding},
  booktitle = {Proceedings of the IEEE/CVF Conference on Computer Vision and Pattern Recognition (CVPR)},
  year      = {2024},
  url       = {https://openaccess.thecvf.com/content/CVPR2024/html/Mu_SnAG_Scalable_and_Accurate_Video_Grounding_CVPR_2024_paper.html}
}

@article{Khurana2023Minotaur,
  author  = {Tarun Khurana and Artem Molchanov and Pavlo Molchanov and Ren Yang and Nagesh Adluru and Jan Kautz and Sachin Raj and Aneeshan Sain},
  title   = {MINOTAUR: Multi-task Video Grounding from Multimodal Queries},
  journal = {arXiv preprint arXiv:2309.13837},
  year    = {2023},
  url     = {https://arxiv.org/abs/2309.13837}
}

@inproceedings{huang2022emb,
 author = {Jiabo Huang and Hailin Jin and Shaogang Gong and Yang Liu},
 booktitle = {Proceedings of the European Conference on Computer Vision (ECCV)},
 title = {Video Activity Localisation with Uncertainties in Temporal Boundary},
 year = {2022}
}

@inproceedings{liu2024towards,
  title={Towards Balanced Alignment: Modal-Enhanced Semantic Modeling for Video Moment Retrieval},
  author={Liu, Zhihang and Li, Jun and Xie, Hongtao and Li, Pandeng and Ge, Jiannan and Liu, Sun-Ao and Jin, Guoqing},
  booktitle={AAAI},
  pages={3855--3863},
  year={2024}
}

@inproceedings{mun2020local,
 author = {Mun, Jonghwan and Cho, Minsu and Han, Bohyung},
 booktitle = {Proceedings of the IEEE/CVF Conference on Computer Vision and Pattern Recognition},
 pages = {10810--10819},
 title = {Local-global video-text interactions for temporal grounding},
 year = {2020}
}

@inproceedings{SPL_2023_ACL,
 author = {Minghang Zheng and Shaogang Gong and Hailin Jin and Yuxin Peng and Yang Liu},
 booktitle = {Annual Meeting of the Association for Computational Linguistics},
 title = {Generating Structured Pseudo Labels for Noise-resistant Zero-shot Video Sentence Localization},
 year = {2023}
}

@inproceedings{VMR_2023_CVPR,
 author = {Dezhao Luo and Jiabo Huang and Shaogang Gong and Hailin Jin and Yang Liu},
 booktitle = {Proceedings of the IEEE/CVF Conference on Computer Vision and Pattern Recognition (CVPR)},
 title = {Towards Generalisable Video Moment Retrieval: Visual-Dynamic Injection to Image-Text Pre-Training},
 year = {2023}
}

@inproceedings{xiao2024bridging,
 author = {Xiao, Yicheng and Luo, Zhuoyan and Liu, Yong and Ma, Yue and Bian, Hengwei and Ji, Yatai and Yang, Yujiu and Li, Xiu},
 booktitle = {Proceedings of the IEEE/CVF Conference on Computer Vision and Pattern Recognition},
 pages = {18709--18719},
 title = {Bridging the gap: A unified video comprehension framework for moment retrieval and highlight detection},
 year = {2024}
}

@inproceedings{zhang2025localizing,
  author    = {Zhang, Gengyuan and Fok, Mang Ling Ada and Ma, Jialu and Xia, Yan and Cremers, Daniel and Torr, Philip and Tresp, Volker and Gu, Jindong},
  title     = {Localizing Events in Videos with Multimodal Queries},
  booktitle = {Proceedings of the IEEE/CVF Conference on Computer Vision and Pattern Recognition (CVPR)},
  pages     = {3339--3351},
  year      = {2025},
  doi       = {10.1109/CVPR52734.2025.00317}
}

@inproceedings{Krishna2017ActivityNetCaptions,
  author    = {Ranjay Krishna and Kenji Hata and Frederic Ren and Li Fei-Fei and Juan Carlos Niebles},
  title     = {Dense-Captioning Events in Videos},
  booktitle = {Proceedings of the IEEE International Conference on Computer Vision (ICCV)},
  year      = {2017},
  url       = {https://openaccess.thecvf.com/content_ICCV_2017/html/Krishna_Dense-Captioning_Events_in_ICCV_2017_paper.html},
}

@inproceedings{shao2020finegym,
title={FineGym: A Hierarchical Video Dataset for Fine-grained Action Understanding},
author={Shao, Dian and Zhao, Yue and Dai, Bo and Lin, Dahua},
booktitle={IEEE Conference on Computer Vision and Pattern Recognition (CVPR)},
year={2020}
}

@misc{wang2024internvidlargescalevideotextdataset,
      title={InternVid: A Large-scale Video-Text Dataset for Multimodal Understanding and Generation}, 
      author={Yi Wang and Yinan He and Yizhuo Li and Kunchang Li and Jiashuo Yu and Xin Ma and Xinhao Li and Guo Chen and Xinyuan Chen and Yaohui Wang and Conghui He and Ping Luo and Ziwei Liu and Yali Wang and Limin Wang and Yu Qiao},
      year={2024},
      eprint={2307.06942},
      archivePrefix={arXiv},
      primaryClass={cs.CV},
      url={https://arxiv.org/abs/2307.06942}, 
}

@inproceedings{Hendricks2017DiDeMo,
  author    = {Lisa Anne Hendricks and Oliver Wang and Eli Shechtman and Josef Sivic and Trevor Darrell and Bryan Russell},
  title     = {Localizing Moments in Video with Natural Language},
  booktitle = {Proceedings of the IEEE International Conference on Computer Vision (ICCV)},
  year      = {2017}
}

@inproceedings{Regneri2013TACoS,
  author    = {Michaela Regneri and Marcus Rohrbach and Dominikus Wetzel and Stefan Thater and Bernt Schiele and Manfred Pinkal},
  title     = {Grounding Action Descriptions in Videos},
  booktitle = {Transactions of the Association for Computational Linguistics (TACL)},
  year      = {2013},
}

@misc{vaswani2023attentionneed,
      title={Attention Is All You Need}, 
      author={Ashish Vaswani and Noam Shazeer and Niki Parmar and Jakob Uszkoreit and Llion Jones and Aidan N. Gomez and Lukasz Kaiser and Illia Polosukhin},
      year={2023},
      eprint={1706.03762},
      archivePrefix={arXiv},
      primaryClass={cs.CL},
      url={https://arxiv.org/abs/1706.03762}, 
}

@article{dong2024temporal,
  title={Temporal sentence grounding with relevance feedback in videos},
  author={Dong, Jianfeng and Peng, Xiaoman and Liu, Daizong and Qu, Xiaoye and Yang, Xun and Bao, Cuizhu and Wang, Meng},
  journal={Advances in Neural Information Processing Systems},
  volume={37},
  pages={43107--43132},
  year={2024}
}

@article{uvcom,
  author       = {Yicheng Xiao and
                  Zhuoyan Luo and
                  Yong Liu and
                  Yue Ma and
                  Hengwei Bian and
                  Yatai Ji and
                  Yujiu Yang and
                  Xiu Li},
  title        = {Bridging the Gap: {A} Unified Video Comprehension Framework for Moment Retrieval and Highlight Detection},
  journal      = {CoRR},
  volume       = {abs/2311.16464},
  year         = {2023}
}

@inproceedings{zhang2022actionformer,
  title={ActionFormer: Localizing Moments of Actions with Transformers},
  author={Zhang, Chen-Lin and Wu, Jianxin and Li, Yin},
  booktitle={European Conference on Computer Vision},
  series={LNCS},
  volume={13664},
  pages={492-510},
  year={2022}
}

@misc{moon2024correlationguidedquerydependencycalibrationvideo,
      title={Correlation-Guided Query-Dependency Calibration for Video Temporal Grounding}, 
      author={WonJun Moon and Sangeek Hyun and SuBeen Lee and Jae-Pil Heo},
      year={2024},
      eprint={2311.08835},
      archivePrefix={arXiv},
      primaryClass={cs.CV},
      url={https://arxiv.org/abs/2311.08835}, 
}

@inproceedings{radford2021clip,
  title={Learning transferable visual models from natural language supervision},
  author={Radford, Alec and Kim, Jong Wook and Hallacy, Chris and Ramesh, Aditya and Goh, Gabriel and Agarwal, Sandhini and Sastry, Girish and Askell, Amanda and Mishkin, Pamela and Clark, Jack and others},
  booktitle={International conference on machine learning},
  pages={8748--8763},
  year={2021},
  organization={PMLR}
}

@article{loshchilov2017adamw,
  title={Decoupled weight decay regularization},
  author={Loshchilov, Ilya and Hutter, Frank},
  journal={arXiv preprint arXiv:1711.05101},
  year={2017}
}

@misc{gemini-2.5-pro,
    title={Gemini 2.5: Pushing the Frontier with Advanced Reasoning, Multimodality, Long Context, and Next Generation Agentic Capabilities},
    author={Google Gemini Team},
    year={2025},
  archivePrefix={arXiv},
  primaryClass={cs.CV},
url={https://arxiv.org/abs/2507.06261}
}

@article{sa2va,
  title={Sa2VA: Marrying SAM2 with LLaVA for Dense Grounded Understanding of Images and Videos},
  author={Yuan, Haobo and Li, Xiangtai and Zhang, Tao and Sun, Yueyi and Huang, Zilong and Xu, Shilin and Ji, Shunping and Tong, Yunhai and Qi, Lu and Feng, Jiashi and Yang, Ming-Hsuan},
  journal={arXiv pre-print},
  year={2025}
}

@article{Qwen3-VL,
      title={Qwen3-VL Technical Report}, 
      author={Shuai Bai and Yuxuan Cai and Ruizhe Chen and Keqin Chen and Xionghui Chen and Zesen Cheng and Lianghao Deng and Wei Ding and Chang Gao and Chunjiang Ge and Wenbin Ge and Zhifang Guo and Qidong Huang and Jie Huang and Fei Huang and Binyuan Hui and Shutong Jiang and Zhaohai Li and Mingsheng Li and Mei Li and Kaixin Li and Zicheng Lin and Junyang Lin and Xuejing Liu and Jiawei Liu and Chenglong Liu and Yang Liu and Dayiheng Liu and Shixuan Liu and Dunjie Lu and Ruilin Luo and Chenxu Lv and Rui Men and Lingchen Meng and Xuancheng Ren and Xingzhang Ren and Sibo Song and Yuchong Sun and Jun Tang and Jianhong Tu and Jianqiang Wan and Peng Wang and Pengfei Wang and Qiuyue Wang and Yuxuan Wang and Tianbao Xie and Yiheng Xu and Haiyang Xu and Jin Xu and Zhibo Yang and Mingkun Yang and Jianxin Yang and An Yang and Bowen Yu and Fei Zhang and Hang Zhang and Xi Zhang and Bo Zheng and Humen Zhong and Jingren Zhou and Fan Zhou and Jing Zhou and Yuanzhi Zhu and Ke Zhu},
	  journal={arXiv preprint arXiv:2511.21631},
      year={2025}
}

@inproceedings{yang2025ar,
  title={Ar-vrm: Imitating human motions for visual robot manipulation with analogical reasoning},
  author={Yang, Dejie and Zhao, Zijing and Liu, Yang},
  booktitle={Proceedings of the IEEE/CVF International Conference on Computer Vision},
  pages={6818--6827},
  year={2025}
}

@inproceedings{mo2025advancing,
  title={Advancing 3D Scene Understanding with MV-ScanQA Multi-View Reasoning Evaluation and TripAlign Pre-training Dataset},
  author={Mo, Wentao and Chen, Qingchao and Peng, Yuxin and Huang, Siyuan and Liu, Yang},
  booktitle={Proceedings of the 33rd ACM International Conference on Multimedia},
  pages={12973--12980},
  year={2025}
}

@InProceedings{blip2,
  title = 	 {{BLIP}-2: Bootstrapping Language-Image Pre-training with Frozen Image Encoders and Large Language Models},
  author =       {Li, Junnan and Li, Dongxu and Savarese, Silvio and Hoi, Steven},
  booktitle = 	 {Proceedings of the 40th International Conference on Machine Learning},
  pages = 	 {19730--19742},
  year = 	 {2023},
  editor = 	 {Krause, Andreas and Brunskill, Emma and Cho, Kyunghyun and Engelhardt, Barbara and Sabato, Sivan and Scarlett, Jonathan},
  volume = 	 {202},
  series = 	 {Proceedings of Machine Learning Research},
  month = 	 {23--29 Jul},
  publisher =    {PMLR},
  url = 	 {https://proceedings.mlr.press/v202/li23q.html},

}

@inproceedings{gao2025omniground,
  title={Omniground: A comprehensive spatio-temporal grounding benchmark for real-world complex scenarios},
  author={Gao, Hong and Wu, Jingyu and Xu, Xiangkai and Xie, Kangni and Zhang, Yunchen and Zhong, Bin and Gao, Xurui and Zhang, Min-Ling},
  booktitle={Proceedings of the IEEE/CVF Conference on Computer Vision and Pattern Recognition},
  pages={17588--17597},
  year={2026}
}

@inproceedings{gao2026sarl,
  title={SARL-STG: A Spatially Aware Reinforcement Learning Framework for Refining MLLMs in Spatio-Temporal Video Grounding},
  author={Gao, Hong and Xu, Xiangkai and Zhong, Bin and Yin, Junjie and Kang, Fangyu and Xu, Yutong and Dong, Xiugang and Gao, Xurui and Zhang, Min-Ling},
  booktitle={Proceedings of the IEEE/CVF Conference on Computer Vision and Pattern Recognition},
  pages={24630--24639},
  year={2026}
}

@article{du2025dataset,
  title={Dataset Copyright Auditing for Large Models: Fundamentals, Open Problems, and Future Directions},
  author={Du, Linkang and Su, Zhou and Yu, Xinyi},
  journal={ZTE COMMUNICATIONS},
  volume={23},
  number={3},
  pages={38--47},
  year={2025},
  publisher={ZTE COMMUNICATIONS}
}

@article{wei2025poison,
  title={Poison-only and targeted backdoor attack against visual object tracking},
  author={GU, Wei and SHAO, Shuo and ZHOU, Lingtao and QIN, Zhan and REN, Kui},
  journal={ZTE COMMUNICATIONS},
  volume={23},
  number={3},
  pages={3--14},
  year={2025},
  publisher={ZTE COMMUNICATIONS}
}

@article{shen2025stegoagent,
  title={StegoAgent StegoAgent: A Generative Steganography A Generative Steganography Framework Based on GUI Agents},
  author={SHEN, Qiuhong and YANG, Zijin and JIANG, Jun and ZHANG, Weiming and CHEN, Kejiang},
  journal={ZTE COMMUNICATIONS},
  volume={23},
  number={3},
  pages={48--58},
  year={2025},
  publisher={ZTE COMMUNICATIONS}
}

@Article{E230419,
title = {Multi-Use Learning Instance for Optimized Image Retrieval},
journal = {Chinese Journal of Electronics},
volume = {34},
number = {3},
pages = {1002-1005},
year = {2025},
author = {Wu Hao and Guo Junqi and Bie Rongfang}
}

@article{yang2025vqals,
  title={VQALS: A Video Question Answering Method in Low-Light Scenes Based on Illumination Correction and Feature Enhancement},
  author={Yang, Jie and Ma, Miao and Li, Yutong and Pei, Zhao},
  journal={Chinese Journal of Electronics},
  volume={34},
  number={4},
  pages={1300--1308},
  year={2025},
  publisher={CIE}
}

@Article{E250336,
title = {A Survey on Fine-Grained Multimodal Large Language Models},
journal = {Chinese Journal of Electronics},
volume = {35},
number = {2},
pages = {771-803},
year = {2026},
issn = {},
author = {Peng Yuxin and Wang Zishuo and Li Geng and Zheng Xiangtian and Yin Sibo and He Hulingxiao}
}

@InProceedings{Zheng_2026_ICML,
    author    = {Zheng, Minghang and Yin, Zihao and Yang, Yi and Peng, Yuxin and Liu, Yang},
    title     = {Temporal-Aware Reasoning Optimization for Video Temporal Grounding},
    booktitle = {International Conference on Machine Learning},
    year      = {2026}
}

@InProceedings{zheng2026omnivtg,
    author    = {Zheng, Minghang and Yin, Zihao and Yang, Yi and Peng, Yuxin and Liu, Yang},
    title     = {OmniVTG: A Large-Scale Dataset and Training Paradigm for Open-World Video Temporal Grounding},
    booktitle = {Proceedings of the IEEE/CVF Conference on Computer Vision and Pattern Recognition (CVPR)},
    month     = {June},
    year      = {2026},
    pages     = {24620-24629}
}

@article{zheng2025weakly,
  title={Weakly and Single-Frame Supervised Temporal Sentence Grounding With Gaussian-Based Contrastive Proposal Learning},
  author={Zheng, Minghang and Huang, Yanjie and Chen, Qingchao and Peng, Yuxin and Liu, Yang},
  journal={IEEE Transactions on Pattern Analysis and Machine Intelligence},
  year={2025},
  publisher={IEEE}
}

@article{liu2026large,
  title={Large-Scale Pre-Trained Models Empowering Phrase Generalization in Temporal Sentence Localization},
  author={Liu, Yang and Zheng, Minghang and Chen, Qingchao and Gong, Shaogang and Peng, Yuxin},
  journal={International Journal of Computer Vision},
  volume={134},
  number={2},
  pages={53},
  year={2026},
  publisher={Springer}
}

@inproceedings{zheng2025hierarchical,
  title={Hierarchical Event Memory for Accurate and Low-latency Online Video Temporal Grounding},
  author={Zheng, Minghang and Peng, Yuxin and Sun, Benyuan and Yang, Yi and Liu, Yang},
  booktitle={Proceedings of the IEEE/CVF International Conference on Computer Vision},
  pages={21589--21599},
  year={2025}
}

\end{document}